\documentclass[onecolumn]{article}
\usepackage{lmodern}
\usepackage{amssymb,amsmath}
\usepackage{ifxetex,ifluatex}
\usepackage{fixltx2e} 
\ifnum 0\ifxetex 1\fi\ifluatex 1\fi=0 
  \usepackage[T1]{fontenc}
  \usepackage[utf8]{inputenc}
\else 
  \ifxetex
    \usepackage{mathspec}
  \else
    \usepackage{fontspec}
  \fi
  \defaultfontfeatures{Ligatures=TeX,Scale=MatchLowercase}
\fi
\IfFileExists{upquote.sty}{\usepackage{upquote}}{}
\IfFileExists{microtype.sty}{%
\usepackage[]{microtype}
\UseMicrotypeSet[protrusion]{basicmath} 
}{}
\PassOptionsToPackage{hyphens}{url} 
\usepackage[unicode=true]{hyperref}
\hypersetup{
            pdftitle={AutoEncoder for Interpolation},
            pdfauthor={Rahul Bhadani},
            pdfborder={0 0 0},
            breaklinks=true}
\usepackage[margin=1in]{geometry}
\usepackage{natbib}
\usepackage{color}
\usepackage{fancyvrb}

\DefineVerbatimEnvironment{Highlighting}{Verbatim}{commandchars=\\\{\}}
\usepackage{framed}
\definecolor{shadecolor}{RGB}{248,248,248}
\newenvironment{Shaded}{\begin{snugshade}}{\end{snugshade}}

\newcommand{\DecValTok}[1]{\textcolor[rgb]{0.00,0.00,0.81}{#1}}

\newcommand{\FloatTok}[1]{\textcolor[rgb]{0.00,0.00,0.81}{#1}}

\newcommand{\StringTok}[1]{\textcolor[rgb]{0.31,0.60,0.02}{#1}}

\newcommand{\ImportTok}[1]{#1}
\newcommand{\CommentTok}[1]{\textcolor[rgb]{0.56,0.35,0.01}{\textit{#1}}}

\newcommand{\VariableTok}[1]{\textcolor[rgb]{0.00,0.00,0.00}{#1}}

\newcommand{\OperatorTok}[1]{\textcolor[rgb]{0.81,0.36,0.00}{\textbf{#1}}}
\newcommand{\BuiltInTok}[1]{#1}

\newcommand{\NormalTok}[1]{#1}
\usepackage{graphicx,grffile}
\makeatletter
\def\maxwidth{\ifdim\Gin@nat@width>\linewidth\linewidth\else\Gin@nat@width\fi}
\def\maxheight{\ifdim\Gin@nat@height>\textheight\textheight\else\Gin@nat@height\fi}
\makeatother
\setkeys{Gin}{width=\maxwidth,height=\maxheight,keepaspectratio}
\IfFileExists{parskip.sty}{%
\usepackage{parskip}
}{
\setlength{\parindent}{0pt}
\setlength{\parskip}{6pt plus 2pt minus 1pt}
}
\ifx\paragraph\undefined\else
\let\oldparagraph\paragraph
\renewcommand{\paragraph}[1]{\oldparagraph{#1}\mbox{}}
\fi
\ifx\subparagraph\undefined\else
\let\oldsubparagraph\subparagraph
\renewcommand{\subparagraph}[1]{\oldsubparagraph{#1}\mbox{}}
\fi

\makeatletter
\def\fps@figure{htbp}
\makeatother

\usepackage[none]{hyphenat}
\graphicspath{ {images/} }
\usepackage{color}
\definecolor{ocre}{RGB}{243,102,25}
\usepackage[font={color=ocre,bf},labelfont=bf]{caption}
\usepackage{hyperref}
\hypersetup{colorlinks=true,citecolor=blue}
\def\BibTeX{{\rm B\kern-.05em{\sc i\kern-.025em b}\kern-.08em T\kern-.1667em\lower.7ex\hbox{E}\kern-.125emX}}

\title{AutoEncoder for Interpolation}
\author{Rahul Bhadani\footnote{The University of Arizona,
  \href{mailto:rahulbhadani@email.arizona.edu}{\nolinkurl{rahulbhadani@email.arizona.edu}}}}
\date{05 January 2021}

\begin{document}
\maketitle
\begin{abstract}
In physical science, sensor data are collected over time to produce
timeseries data. However, depending on the real-world condition and
underlying physics of the sensor, data might be noisy. Besides, the
limitation of sample-time on sensors may not allow collecting data over
all the timepoints, may require some form of interpolation.
Interpolation may not be smooth enough, fail to denoise data, and
derivative operation on noisy sensor data may be poor that do not reveal
any high order dynamics. In this article, we propose to use Autoencoder
to perform interpolation that also denoise data simultaneously. A brief
example using a real-world is also provided.
\end{abstract}

\section{Autoencoder}\label{autoencoder}

\begin{figure}
\centering
\includegraphics[width=0.50000\textwidth]{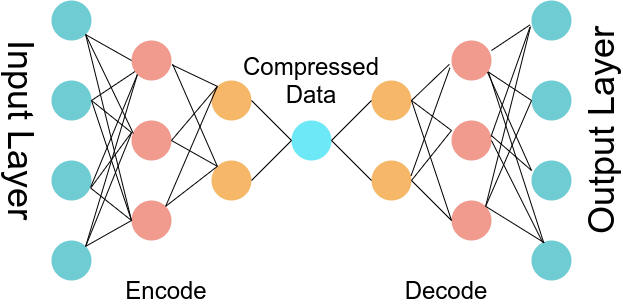}
\caption{A Schematic of Autoencoder drawn using
draw.io\label{neuralnet}}
\end{figure}

Autoencoders \citep{ballard1987modular} are another Neural Network used
to reproduce the inputs in a compressed fashion. Autoencoder has a
special property in which the number of input neurons is the same as the
number of output neurons. See the Figure \ref{neuralnet}. The goal of
Autoencoder is to create a representation of the input at the output
layer such that both output and input are similar but the actual use of
the Autoencoder is for determining a compressed version of the input
data with the lowest amount of loss in data. This is very similar to
what Principal Component Analysis does, in a black-box manner. Encoder
part of Autoencoder compresses the data at the same time ensuring that
the important data is not lost but the size of the data is reduced.

The downside of using Autoencoder for interpolation is that the
compressed data is a black box representation--- we do not know the
structure of the data in the compressed version. Suppose we have a
dataset with 10 parameters and we train an autoencoder over this data.
The encoder does not omit some of the parameters for better
representation but it fuses the parameters to create a compressed
version but with fewer parameters (brings the number of parameters down
to, say, 5 from 10). Autoencoder has two parts, encoder, and decoder.
The encoder compresses the input data and the decoder does the opposite
to produce the uncompressed version of the data to produce a
reconstructed input as close to the original one as possible.

\section{Interpolation Method}\label{interpolation-method}

\begin{figure}
\centering
\includegraphics[width=0.80000\textwidth]{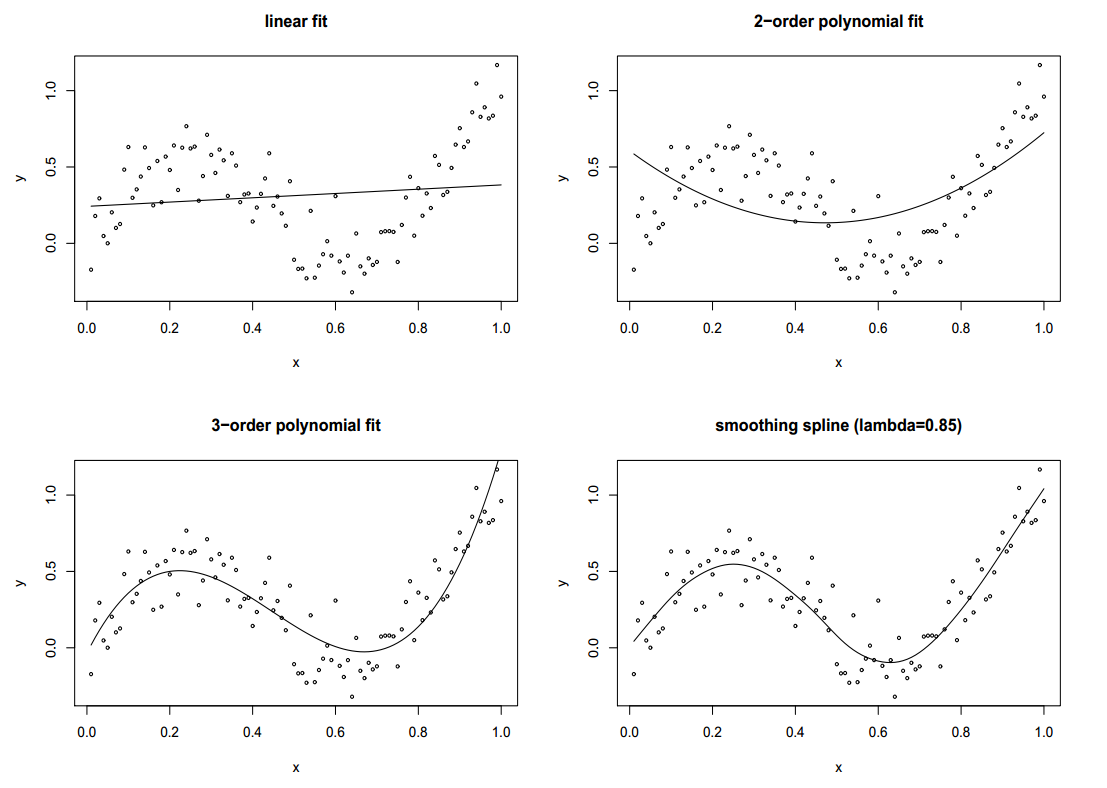}
\caption{Examples of data-fitting.\label{datafitting}}
\end{figure}

Interpolation is a process of guessing the value of a function between
two data points. For example, you are given \(x = [1, 3, 5, 7, 9]\), and
\(y = [230.02, 321.01, 305.00, 245.75, 345.62]\), and based on the given
data you want to know the value of \(y\) given \(x = 4\). There are
plenty of interpolation methods available in the literature --- some
model-based and some are model-free, i.e.~data-driven. The most common
way of achieving interpolation is through data-fitting. As an example,
you use linear regression analysis to fit a linear model to the given
data. In linear regression \citep{kutner2005applied}, given the
explanatory/predictor variable, \(X\), and the response variable, \(Y\),
the data is fitted using the formula \(Y = \beta_0 + \beta_1 X\) where
\(\beta_0\) and \(\beta_1\) are determined using least square fit. As
the name suggests, linear regression is linear, i.e., it fits a straight
line even though the relationship between predictor and response
variable might be non-linear. However, the most general form of
interpolation is polynomial fitting. Given k sample points, it is
straightforward to fit a polynomial of degree \(k -1\). Given the data
set \(\{x_i, y_i\}\), the polynomial fitting is obtained by determining
polynomial coefficients \(a_i\) of function

\[
f(x) = a_0 + a_1 x  + a_2 x^2 + \cdots + a_{k-1}x^{k-1}
\]

by solving matrix inversion from the following expression:

\[
\begin{pmatrix}1 & x_1 & x_1^2 & \cdots & x_1^{k-1}\\1 & x_2 & x_2^2 & \cdots & x_2^{k-1} \\ \vdots & \vdots & \vdots & \cdots & \vdots \\ 1 & x_1 & x_1^2 & \cdots & x_1^{k-1}
\end{pmatrix}
\begin{pmatrix}
a_0\\a_1 \\ \vdots \\ a_{k-1}
\end{pmatrix} = \begin{pmatrix} y_0 \\ y_1 \\ \vdots \\ y_{k-1}\end{pmatrix}
\] Once we have coefficients \(a_i\), we can find the value of function
\(f\) for any \(x\). There are some specific cases of polynomial fitting
where a piecewise cubic polynomial is fitted to data. A few other
non-parametric methods include cubic spline, smoothing splines,
regression splines, kernel regression, and density estimation
\citep{hastie2009elements} (Figure \ref{datafitting}).

However, the point of this article is not polynomial fitting, but rather
interpolation. Polynomial fitting just happens to facilitate
interpolation. However, there is an issue with polynomial fitting
methods --- whether it is parametric or non-parametric, they behave the
way they are taught. What it means is that if data is clean, the fitting
will be clean and smooth, but if data is noisy, the fitting will be
noisy. This issue is more prevalent in sensor data, for example,
hear-beat data captured from your heart-rate sensor, distance data from
LiDAR, CAN Bus speed data from your car, GPS data, etc.

Further, because of the noise, they are harder to deal with, especially
if your algorithm requires performing double, or second derivative on
such data. In general, those sensor data are timeseries data, i.e.~they
are collected over time, thus the response variable might be some
physical quantity such as speed, the distance of objects from LiDAR
mounted on the top of a self-driving car, heart-rate, and predictor
variable is time. While operating on such data, there can be a few
objectives: we want to have data interpolated to some time-stamp over
which our sensor couldn't record any response, but since sensors operate
in the real-time world and because of the underlying physics, those data
stay noisy, we also want a reliable interpolation that is not impacted
by sensor noise. Further, our requirement may also include derivatives
of such timeseries data. Derivatives tend to amplify the noise present
in the underlying timeseries data \citep{bhadani2019real}. What if there
is a way by which we can get an underlying representation of the data,
discarding the noise at the same time? Autoencoder comes to the rescue
to achieve our objective in such a case.

\section{Autoencoder as Interpolator}\label{autoencoder-as-interpolator}

To demonstrate the denoising + interpolation objective using
Autoencoder, we use an example of distance data collected from a vehicle
by our lab, where the response variable is the distance of the vehicle
ahead of our vehicle, and the predictor is time. We have made a small
subset of the data available on the GitHub repo at
\url{https://github.com/rahulbhadani/medium.com/blob/master/data/lead_dist_sample.csv}
as a part of the demonstration that you are free to use. However, it is
really small and serves no purpose beyond the tutorial described in this
article.

\textbf{Note:} \emph{Before you use data, we should point out that the
time (predictor) and message (response) must be re-scaled. In our case,
the original time starts from 1594247088.289515 (in POSIX format, in
seconds) and ends at 1594247110.290019. We normalized the time value
using the formula
\texttt{(time\ -\ start\_time)/(end\_time\ -\ start\_time)}. Similarly,
the response variable was normalized using
\texttt{(message\ -\ message\_min)/(message\_max\ -message\_min)}. The
sample data provided in the GitHub repo is already normalized and you
can reuse it out of the box.}

\subsection{Training}\label{training}

\begin{Shaded}
\begin{Highlighting}[]
\ImportTok{import}\NormalTok{ glob}
\ImportTok{import}\NormalTok{ numpy }\ImportTok{as}\NormalTok{ np}
\ImportTok{import}\NormalTok{ matplotlib.pyplot }\ImportTok{as}\NormalTok{ plt}
\ImportTok{import}\NormalTok{ pandas }\ImportTok{as}\NormalTok{ pd}
\ImportTok{import}\NormalTok{ numpy }\ImportTok{as}\NormalTok{ np}
\NormalTok{df }\OperatorTok{=}\NormalTok{ pd.read_csv(}\StringTok{"../data/lead_dist_sample.csv"}\NormalTok{)}
\NormalTok{time }\OperatorTok{=}\NormalTok{ df[}\StringTok{'Time'}\NormalTok{]}
\NormalTok{message }\OperatorTok{=}\NormalTok{ df[}\StringTok{'Message'}\NormalTok{]}
\ImportTok{import}\NormalTok{ tensorflow }\ImportTok{as}\NormalTok{ tf}
\NormalTok{model }\OperatorTok{=}\NormalTok{ tf.keras.Sequential()}
\NormalTok{model.add(tf.keras.layers.Dense(units }\OperatorTok{=} \DecValTok{1}\NormalTok{, activation }\OperatorTok{=} \StringTok{'linear'}\NormalTok{, input_shape}\OperatorTok{=}\NormalTok{[}\DecValTok{1}\NormalTok{]))}
\NormalTok{model.add(tf.keras.layers.Dense(units }\OperatorTok{=} \DecValTok{128}\NormalTok{, activation }\OperatorTok{=} \StringTok{'relu'}\NormalTok{))}
\NormalTok{model.add(tf.keras.layers.Dense(units }\OperatorTok{=} \DecValTok{64}\NormalTok{, activation }\OperatorTok{=} \StringTok{'relu'}\NormalTok{))}
\NormalTok{model.add(tf.keras.layers.Dense(units }\OperatorTok{=} \DecValTok{32}\NormalTok{, activation }\OperatorTok{=} \StringTok{'relu'}\NormalTok{))}
\NormalTok{model.add(tf.keras.layers.Dense(units }\OperatorTok{=} \DecValTok{64}\NormalTok{, activation }\OperatorTok{=} \StringTok{'relu'}\NormalTok{))}
\NormalTok{model.add(tf.keras.layers.Dense(units }\OperatorTok{=} \DecValTok{128}\NormalTok{, activation }\OperatorTok{=} \StringTok{'relu'}\NormalTok{))}
\NormalTok{model.add(tf.keras.layers.Dense(units }\OperatorTok{=} \DecValTok{1}\NormalTok{, activation }\OperatorTok{=} \StringTok{'linear'}\NormalTok{))}
\NormalTok{model.}\BuiltInTok{compile}\NormalTok{(loss}\OperatorTok{=}\StringTok{'mse'}\NormalTok{, optimizer}\OperatorTok{=}\StringTok{"adam"}\NormalTok{)}
\NormalTok{model.summary()}
\CommentTok{# Training}
\NormalTok{model.fit( time, message, epochs}\OperatorTok{=}\DecValTok{1000}\NormalTok{, verbose}\OperatorTok{=}\VariableTok{True}\NormalTok{)}
\end{Highlighting}
\end{Shaded}

As you can see, We have not performed any regularization as we
deliberately want to do overfitting so that we can use the underlying
nature of data to the full extent. Now it's time to make a prediction.
You will see that we rescaled back the time axis to original values
before making predictions. For this example, we had
\texttt{time\_original{[}0{]}\ =\ 1594247088.289515}
,\texttt{time\_original{[}-1{]}\ =\ 1594247110.290019} ,
\texttt{msg\_min\ =\ 33}, \texttt{msg\_max\ =\ 112}.

\begin{Shaded}
\begin{Highlighting}[]
\NormalTok{newtimepoints_scaled }\OperatorTok{=}\NormalTok{ np.linspace(time[}\DecValTok{0}\NormalTok{] }\OperatorTok{-}\NormalTok{ (time[}\DecValTok{1}\NormalTok{] }\OperatorTok{-}\NormalTok{ time[}\DecValTok{0}\NormalTok{]),time[}\OperatorTok{-}\DecValTok{1}\NormalTok{], }\DecValTok{10000}\NormalTok{)}
\NormalTok{y_predicted_scaled }\OperatorTok{=}\NormalTok{ model.predict(newtimepoints_scaled)}
\NormalTok{newtimepoints }\OperatorTok{=} 
\NormalTok{    newtimepoints_scaled}\OperatorTok{*}\NormalTok{(time_original[}\OperatorTok{-}\DecValTok{1}\NormalTok{] }\OperatorTok{-}\NormalTok{ time_original[}\DecValTok{0}\NormalTok{]) }\OperatorTok{+}\NormalTok{ time_original[}\DecValTok{0}\NormalTok{]}
\NormalTok{y_predicted }\OperatorTok{=}\NormalTok{ y_predicted_scaled}\OperatorTok{*}\NormalTok{(msg_max }\OperatorTok{-}\NormalTok{ msg_min) }\OperatorTok{+}\NormalTok{ msg_min}
\end{Highlighting}
\end{Shaded}

\begin{figure}
\centering
\includegraphics[width=1.00000\textwidth]{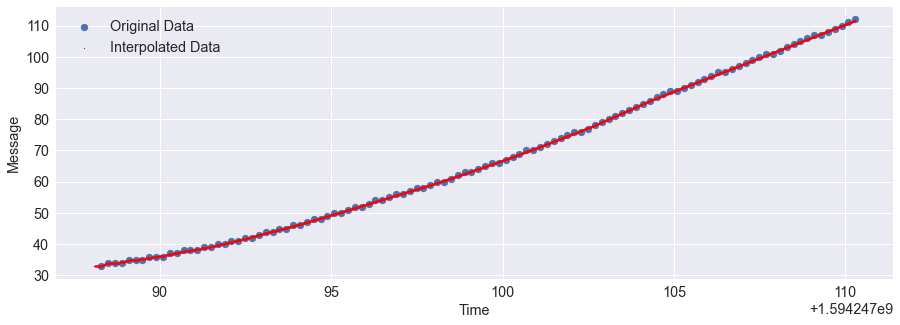}
\caption{Interpolated Data and Original Data\label{interpolated}}
\end{figure}

Note that we are creating much denser time-points in variable
\texttt{newtimepoints\_scaled} which allows me to interpolate data on
unseen time-points. Finally, the is the curve comparing interpolated
data and original data is shown in Figure \ref{interpolated}.

\section{Concluding Remarks}\label{concluding-remarks}

\begin{figure}
\centering
\includegraphics[width=1.00000\textwidth]{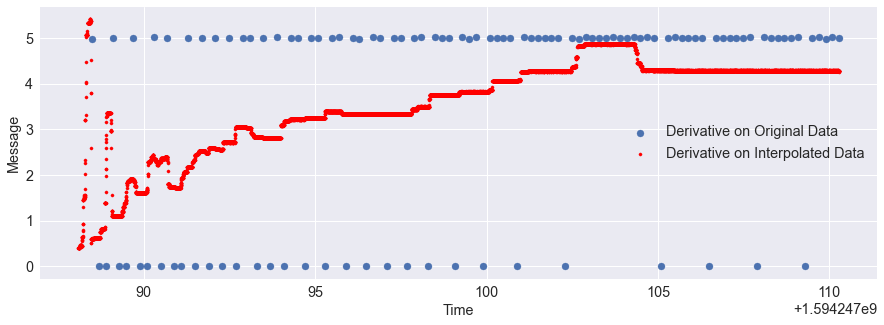}
\caption{A comparison of calculating derivative on original and
interpolated data\label{derivatives}}
\end{figure}

While we trained for only 1000 epochs, your training might not be that
short, if your data is big. The biggest advantage of this method is
taking derivatives, as from Figure \ref{derivatives}, it is clear that
the derivative performed on the original data is poor --- may not even
represent the true derivative!

\begin{Shaded}
\begin{Highlighting}[]
\NormalTok{df_interpolation }\OperatorTok{=}\NormalTok{ pd.DataFrame()}
\NormalTok{df_interpolation[}\StringTok{'Time'}\NormalTok{] }\OperatorTok{=}\NormalTok{ newtimepoints}
\NormalTok{df_interpolation[}\StringTok{'Message'}\NormalTok{] }\OperatorTok{=}\NormalTok{ y_predicted}
\NormalTok{df_interpolation[}\StringTok{'diff'}\NormalTok{] }\OperatorTok{=}
\NormalTok{    df_interpolation[}\StringTok{'Message'}\NormalTok{].diff()}\OperatorTok{/}\NormalTok{df_interpolation[}\StringTok{'Time'}\NormalTok{].diff()}
\NormalTok{df_original }\OperatorTok{=}\NormalTok{ pd.DataFrame()}
\NormalTok{df_original[}\StringTok{'Time'}\NormalTok{] }\OperatorTok{=}\NormalTok{ time}\OperatorTok{*}\NormalTok{(}\FloatTok{1594247110.290019} \OperatorTok{-} \FloatTok{1594247088.289515}\NormalTok{) }\OperatorTok{+} \FloatTok{1594247088.289515}
\NormalTok{df_original[}\StringTok{'Message'}\NormalTok{]  }\OperatorTok{=}\NormalTok{ message}\OperatorTok{*}\NormalTok{(}\DecValTok{112} \OperatorTok{-} \DecValTok{33}\NormalTok{) }\OperatorTok{+} \DecValTok{33}
\NormalTok{df_original[}\StringTok{'diff'}\NormalTok{] }\OperatorTok{=}\NormalTok{ df_original[}\StringTok{'Message'}\NormalTok{].diff()}\OperatorTok{/}\NormalTok{df_original[}\StringTok{'Time'}\NormalTok{].diff()}
\CommentTok{# Display the result}
\ImportTok{import}\NormalTok{ matplotlib.pylab }\ImportTok{as}\NormalTok{ pylab}
\NormalTok{params }\OperatorTok{=}\NormalTok{ \{}\StringTok{'legend.fontsize'}\NormalTok{: }\StringTok{'x-large'}\NormalTok{,}
          \StringTok{'figure.figsize'}\NormalTok{: (}\DecValTok{15}\NormalTok{, }\DecValTok{5}\NormalTok{),}
         \StringTok{'axes.labelsize'}\NormalTok{: }\StringTok{'x-large'}\NormalTok{,}
         \StringTok{'axes.titlesize'}\NormalTok{:}\StringTok{'x-large'}\NormalTok{,}
         \StringTok{'xtick.labelsize'}\NormalTok{:}\StringTok{'x-large'}\NormalTok{,}
         \StringTok{'ytick.labelsize'}\NormalTok{:}\StringTok{'x-large'}\NormalTok{\}}
\NormalTok{pylab.rcParams.update(params)}
\NormalTok{plt.scatter(df_original[}\StringTok{'Time'}\NormalTok{], df_original[}\StringTok{'diff'}\NormalTok{], label}\OperatorTok{=}\StringTok{'Derivative on Original Data'}\NormalTok{)}
\NormalTok{plt.scatter(df_interpolation[}\StringTok{'Time'}\NormalTok{], df_interpolation[}\StringTok{'diff'}\NormalTok{], }
\NormalTok{    s}\OperatorTok{=} \DecValTok{10}\NormalTok{, c }\OperatorTok{=} \StringTok{'red'}\NormalTok{, label}\OperatorTok{=}\StringTok{'Derivative on Interpolated Data'}\NormalTok{)}
\NormalTok{plt.xlabel(}\StringTok{'Time'}\NormalTok{)}
\NormalTok{plt.ylabel(}\StringTok{'Message'}\NormalTok{)}
\NormalTok{plt.legend()}
\NormalTok{plt.show()}
\end{Highlighting}
\end{Shaded}

\section{Acknowledgement}\label{acknowledgement}

Originally, this article was produced by the author in Medium.com at
\url{https://rahulbhadani.medium.com/tensorflow-2-how-to-use-autoencoder-for-interpolation-91fefd4516c9}.

\renewcommand\refname{References}
\bibliography{biblio.bib}

\end{document}